\documentclass[10pt,twocolumn,letterpaper]{article}

\usepackage{cvpr}
\usepackage{times}
\usepackage{epsfig}
\usepackage{graphicx}
\usepackage{amsmath}
\usepackage{amssymb}
\usepackage{multirow}
\usepackage{xcolor}

\usepackage[breaklinks=true,bookmarks=false]{hyperref}

\cvprfinalcopy 

\def\cvprPaperID{****} 

\begin{document}

\title{Attentive Normalization for Conditional Image Generation}

\author{Yi Wang$^{1*}$ \quad Ying-Cong Chen$^{1}$ \quad Xiangyu Zhang$^{2}$ \quad Jian Sun$^{2}$ \quad Jiaya Jia$^{1}$\\
	$^{1}$The Chinese University of Hong Kong \quad $^{2}$MEGVII Technology\\
	{\tt\small \{yiwang, ycchen, leojia\}@cse.cuhk.edu.hk \quad \{zhangxiangyu, sunjian\}@megvii.com}
}

\maketitle

\begin{abstract}
	
	Traditional convolution-based generative adversarial networks synthesize images based on hierarchical local operations, where long-range dependency relation is implicitly modeled with a Markov chain. It is still not sufficient for categories with complicated structures. 
	In this paper, we characterize long-range dependence with attentive normalization (AN), which is an extension to traditional instance normalization. 
	Specifically, the input feature map is softly divided into several regions based on its internal semantic similarity, which are respectively normalized. 
	It enhances consistency between distant regions with semantic correspondence. 
	Compared with self-attention GAN, our attentive normalization does not need to measure the correlation of all locations, and thus can be directly applied to large-size feature maps without much computational burden. Extensive experiments on class-conditional image generation and semantic inpainting verify the efficacy of our proposed module.
	
\end{abstract}

\makeatletter
\def\blfootnote{\gdef\@thefnmark{}\@footnotetext}
\makeatother

\blfootnote{*Part of the work is formed when YW took an internship at MEGVII Technology. \\The research of Zhang and Sun is supported by The National Key Research and Development Program of China (No. 2017YFA0700800) and Beijing Academy of Artificial Intelligence (BAAI). }

\section{Introduction}

Generative adversarial networks \cite{goodfellow2014generative} make image generation attract much attention. It aims to generate realistic images based on a collection of natural images.  
This allows various practical applications, \eg\ image creation \cite{karras2018style,karras2017progressive}, editing \cite{yu2018generative,wang2018inpainting}, data augmentation in discriminative tasks \cite{antoniou2017data}, etc. 

\begin{figure}
	\centering
	\includegraphics[width=1\linewidth]{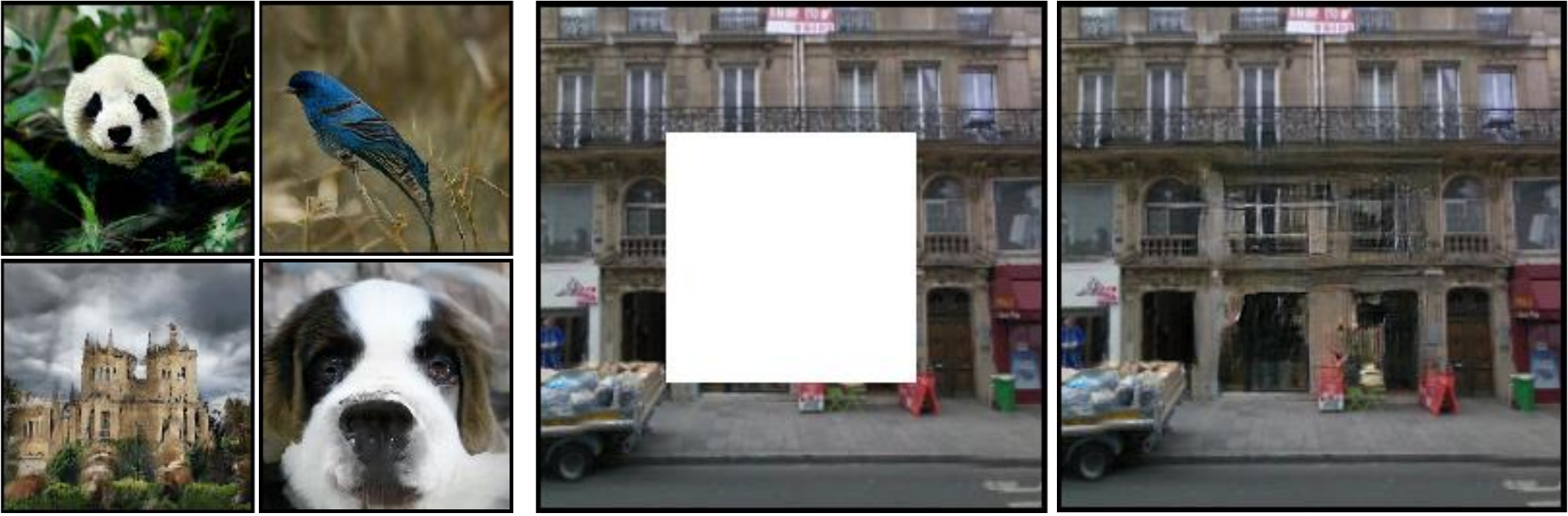}
	\begin{tabular}{cc}
		\hspace{-0.16\columnwidth}(a) & \hspace{0.4\columnwidth}(b) \\
	\end{tabular}
	\caption{Conditional image generation of a GAN framework using our proposed attentive normalization module. (a) Class-conditional image generation. (b) Image inpainting.}
	\label{fig:teaser}
	\vspace{-0.1in}
\end{figure}

Most image generators rely on a fully convolutional generator \cite{radford2015unsupervised,salimans2016improved,miyato2018cgans}. Although these approaches have demonstrated their success in modeling structured data like human faces \cite{karras2017progressive,karras2018style}, and unstructured data such as natural scene \cite{miyato2018cgans,miyato2018spectral}, they do not work well on complicated structured data such as cats or dogs. The reason is that each layer in convolutional neural networks (CNN) is locally bounded, and the relation between distant locations relies on the Markovian modeling between convolutional layers. 

In this regard, although stacking convolution layers could lead to the large receptive field, fully convolutional generators still lack the power to model high-order relationship in distant locations. Such long-range relation is vital because it presents the semantic correspondence that human perception is familiar with and sensitive about, \eg\ symmetry of natural objects and correspondence among limbs.

Self-attention GAN (SA-GAN) \cite{zhang2018self} takes the first step to model long-range dependency in class-conditional image generation.
It introduces a self-attention module in the convolution-based generator, which is helpful for capturing the relation of distant regions. However, the self-attention module requires computing the correlation between every two points in the feature map. Therefore, the computational cost grows rapidly as the feature map becomes large. 
In this paper, we propose a different way for long-range dependency modeling, and achieves better results as well as a lower computational burden. 

Our method is built upon instance normalization (IN). But the previous solution of (IN) normalizes the mean and variance of a feature map along its spatial dimensions. This strategy ignores the fact that different locations may correspond to semantics with varying mean and variance. As illustrated in \cite{park2019semantic}, this mechanism tends to deteriorate the learned semantics of the intermediate features spatially.
 
In this paper, we normalize the input feature maps spatially according to the semantic layouts predicted from them. It improves the distant relationship in the input as well as preserving semantics spatially. In our method, estimation of the semantic layouts relies on two empirical observations. 
First, a feature map can be viewed as a composition of multiple semantic entities \cite{greff2017neural}. Second, the deep layers in a neural network capture high-level semantics of the input images \cite{Quoc2012unsupervised}.
 
We propose our semantic layout learning module based on these observations. This module contains two components, \ie, semantic layout prediction, and self-sampling regularization. The former produces semantic-aware masks that divide the feature map into several parts. Self-sampling regularization regularizes optimization of semantic layout prediction, avoiding trivial results. 

With the semantic layout, spatial information propagation is conducted by the independent normalization in each region. This naturally enhances the relationship between feature points with similar semantics beyond the spatial limit, because their distribution becomes compact via normalization. Their common characteristics are preserved and even enhanced through their exclusive learnable affine transformation. 

The proposed normalization is general. It is experimentally validated in the class conditional image generation (on ImageNet \cite{deng2009imagenet}) and generative image inpainting (on Paris Streetview \cite{pathak2016context}). Figure \ref{fig:teaser} shows a few results. Our major contribution is the following. 
\begin{itemize}
	\vspace{-0.1in}
	\item We propose an attentive normalization (AN) to capture visual distant relationship in the intermediate feature maps during image generation. AN predicts a semantic layout from the input feature map and then conduct regional instance normalization on the feature map based on this layout. 
	\vspace{-0.1in}
	\item The proposed AN module has a low computation complexity by simultaneously fusing and propagating feature statistics in regions with similar semantics.
	\vspace{-0.1in}
	\item Extensive experiments are conducted to prove the effectiveness of AN in distant relationship modeling on class-conditional image generation and generative image inpainting. With the same or similar training setting and model capacity, the proposed AN module achieves comparable or superior visual and quantitative results. In the class conditional image generation task on ImageNet ($128 \times 128$), Frechet Inception Distance (FID) \cite{heusel2017gans} reaches 17.84, compared with 18.65 achieved by self-attention GAN \cite{zhang2018self}, and 22.96 without these long-range dependency modeling modules.
\end{itemize}
\section{Related Work}

\subsection{Generative Adversarial Networks}
The generative adversarial network (GAN) \cite{goodfellow2014generative} is an effective model to synthesize new images, by learning to map random noise to real image samples. 
However, GAN training is usually difficult considering its sensitivity to the model design and parameters. A lot of methods were proposed to improve the procedure, including the architecture design for the generator and discriminator \cite{radford2015unsupervised,karras2017progressive,karras2018style,miyato2018cgans}, more stable distribution measurement for learning objective \cite{mao2017least,arjovsky2017wasserstein,jolicoeur2018relativistic}, model weight and gradients constraints \cite{gulrajani2017improved,miyato2018spectral}, to name a few.


\subsection{Attention in Long Range Dependency Modeling}

Attention modules in neural networks explicitly model the relation between neural elements based on their correlation, serving as a crucial component in various natural language processing and computer vision tasks. In image generation, distant relationship modeling via attention mechanisms is proved to be effective for learning high-dimensional and complex image distribution \cite{zhang2018self,xu2018attngan,fu2019dual,huang2019ccnet,huang2019interlaced}. 

In \cite{zhang2018self}, the proposed self-attention module reconstructs each feature point using the weighted sum of all feature points. This module significantly improves the correlation between distant relevant regions in the feature map, showing obvious advances in large-scale image generation. From the computation perspective, pair-wise relationship calculation in the feature map demands quadratic complexity (regarding both time and space), limiting its application to large feature maps.  

\subsection{Normalization in Deep Learning}

Normalization is vital in neural network training regarding both discriminative or generative tasks. It makes the input features approach independent and identical distribution by a shared mean and variance. This property accelerates training convergence of neural networks and makes training deep networks feasible. Practical normalization layers include batch normalization \cite{ioffe2015batch}, instance normalization \cite{ulyanov2016instance}, layer normalization \cite{ba2016layer}, and group normalization \cite{wu2018group}, which are common in deep learning based classifiers. 

Besides, some normalization variants find applications in image generation tasks with additional conditions, \eg\ conditional batch normalization (CBN) \cite{miyato2018cgans}, adaptive instance normalization (AdaIN) \cite{huang2017arbitrary}, and spatially-adaptive (de)normalization (SPADE) \cite{park2019semantic}. Generally, after normalizing the given feature maps, these features are further affine-transformed, which is learned upon other features or conditions. These ways of conditional normalization can benefit the generator in creating more plausible label-relevant content.

\begin{figure}[!t]
	\centering
	\includegraphics[width=1\linewidth]{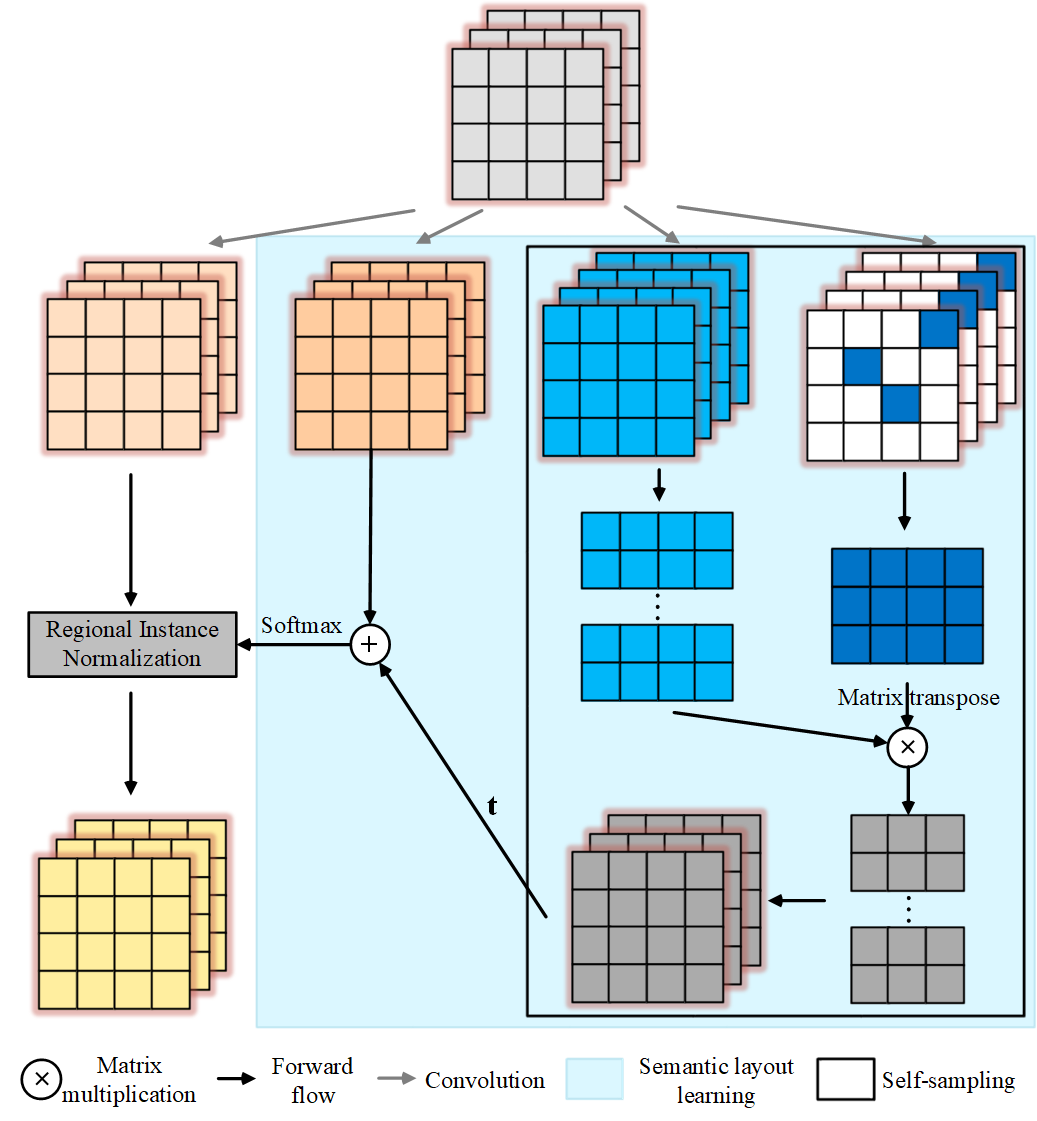}
	\vspace{-0.1in}
	\caption{Proposed attentive normalization module.}
	\label{fig:attentive-normalization-frame}
	\vspace{-0.1in}
\end{figure}

\section{Attentive Normalization} \label{sec_an}

The idea of Attentive Normalization (AN) is to divide the feature maps into different regions based on their semantics, and then separately normalize and de-normalize the feature points in the same region. The first task is addressed by the proposed semantic layout learning  (SLL) module, and the second one is conducted by the regional normalization.

For the given feature maps $\mathbf{X} \in \mathcal{R}^{h \times w \times c}$, Attentive Normalization (AN) learns a soft semantic layout $\mathbf{L} \in \mathcal{R}^{h \times w \times n}$ and normalizes $\mathbf{X}$ spatially according to $\mathbf{L}$, where $\mathbf{L}_p \in [0, 1]$, $n$ denotes a predefined class number, and $p$ denotes pixel location. 

AN is formed by the proposed semantic layout learning  (SLL) module, and a regional normalization, as shown in Figure \ref{fig:attentive-normalization-frame}. It has a semantics learning branch and a self-sampling branch. The semantic learning branch employs a certain number of convolutional filters to capture regions with different semantics (which are activated by a specific filter), with the assumption that each filter in this branch corresponds to some semantic entities. 

The self-sampling branch is complementary to the former semantic learning one. It regularizes learning of the semantic entities so that the semantic learning branch can avoid producing useless semantics -- it means they are uncorrelated to the input features. Combining the output from these two branches, the layout is computed via softmax. Then the regional normalization is conducted on an affine translated feature maps according to such layout.

\subsection{Semantic Layout Learning Module} \label{sec_sll}
We assume each image is composed of $n$ semantic entities. For each feature point from the feature map of the image, it is determined by at least one entity. This assumption gives an expressive representation since these entities can be employed to known novel objects in different contexts. Such assumptions were widely used in unsupervised representation learning \cite{Quoc2012unsupervised}. 

Here we are interested in the way to group feature points of an image according to their correlation to the semantic entities. It helps enhance intra-similarity in the same group. We give $n$ initial desired semantic entities, and define their correlation to the feature points of the image as their inner product. The semantics to represent these entities are learned through back-propagation. We aggregate the feature points from the input feature maps into different regions based on the activation status with these entities. 

Further, to encourage these entities to approach diverse patterns, orthogonal regularization is employed to these entities as
\begin{equation}
\mathcal{L}_{o}=\lambda_{o}||\mathbf{W} \mathbf{W}^\text{T}-\mathbf{I}||^2_\text{F},
\end{equation}
where $\mathbf{W} \in \mathcal{R}^{n \times c}$ is a weight matrix constituted by these $n$ entities (each row is the spanned weight in the row-vector form).

In our implementation, a convolutional layer with $n$ filters is adopted as semantic entities. This layer transforms the input feature maps $\mathbf{X}$ into new feature space as $f(\mathbf{X}) \in \mathcal{R}^{h \times w \times n}$. Intuitively, the larger $n$ is, the more diverse and rich high-level features can be learned. $n=16$ is empirically good for conduct $128 \times 128$ class-conditional image generation and $256 \times 256$ generative image inpainting.  

However, only relying on this component does not lead to reasonable training since it tends to group all feature points with a single semantic entity. It is caused by not setting protocols to ban useless semantic entities that have low or no correlation with the input feature points. From this perspective, we introduce a self-sampling branch providing a reasonable initial semantic layout estimate. It can prevent the trivial solution.

\vspace{-0.05in}
\paragraph{Self-sampling Regularization}
Besides learning the aforementioned semantic layout from scratch, we regularize semantics learning with a self-sampling branch. It is inspired by the practice in feature quantization \cite{xu2005maximum,caron2018deep}, which reassigns empty clusters with the centroid of a non-empty cluster. 

Our self-sampling branch randomly selects $n$ feature points from the input translated feature maps, acting as the alternatives for the semantic entities. They are activated when some entities become irrelevant with the input feature maps. This branch utilizes the correlations in the same feature maps to approximate the semantic layout.

Specifically, this branch randomly (we use uniform sampling) selects $n$ feature pixels from the translated feature maps $k(\mathbf{X})$ as initial semantic filters. To capture more salient semantics, $k(\mathbf{X})$ is processed by max-pooling first.
Then an activation status map $\mathbf{F}$ is calculated as
\begin{equation}
\mathbf{F}_{i,j} = k(\mathbf{X})_i^\text{T}q(\mathbf{X})_j,
\end{equation}
where $\mathbf{F} \in \mathcal{R}^{h \times w \times n}$. $q(\mathbf{X})$ are also translated feature maps. $i$ and $j$ denote pixel location. We set $\#\{i\}=n$ and $\#\{j\}=h\times w$.

\subsection{Soft Semantic Layout Computation}
With the slowly updated $f(\mathbf{X})$ and fast generated $\mathbf{F}$, the raw semantics activation maps $\mathbf{S}^\text{raw}$ are computed as
\begin{equation}
\mathbf{S}^\text{raw} = \mathbf{t} \mathbf{F} + f(\mathbf{X}),
\end{equation}
where $\mathbf{t} \in \mathcal{R}^{1 \times 1 \times n}$ is a learnable vector initialized as $\mathbf{0.1}$. It adaptively adjusts the influence of the self-sampling branch, making the self-sampling branch offer meaningful entity alternatives when some entities become useless during training.

Then we normalize $\mathbf{S}^\text{raw}$ using softmax to get the soft semantic layout as
\begin{equation}
\mathbf{S}_k = \frac{\exp (\tau \mathbf{S}^\text{raw}_k)}{\sum_{i=1}^{n} \exp (\tau \mathbf{S}^\text{raw}_i)},
\end{equation}
where $i$ and $k$ index the feature channels. Each $\mathbf{S}_k$ is a soft mask, indicating the probability of every pixel belonging to class $k$. $\tau$ is the coefficient to control the smoothness of the predicted semantic layout with default value set to $0.1$.

\begin{figure}
	\centering
	\includegraphics[width=0.5\linewidth]{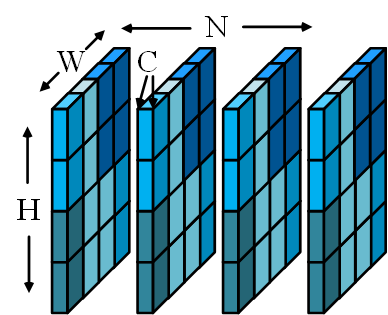}
	\caption{Illustration of regional normalization. The shown feature maps are segmented into four different regions (each with a color) spatially. Each mean and variance are computed on the feature points of the same color in every feature map. $N$, $H$, $W$, and $C$ denote the batch size, channel number, height, and width, respectively.}
	\label{fig:regional-norm}
	\vspace{-0.05in}
\end{figure}

\subsection{Regional Normalization} 
With the soft semantic layout, long-range relationship in feature maps is modeled by regional instance normalization. It considers spatial information and treats each individual region as an instance (shown in Figure \ref{fig:regional-norm}). Correlation between feature points with the same or similar semantics are improved through shared mean and variance, as
\begin{equation}
\mathbf{\bar{X}} = \sum_{i=1}^{n} (\frac{\mathbf{X}-\mu(\mathbf{X}_{\mathbf{S}_i})}{\sigma(\mathbf{X}_{\mathbf{S}_i})+\epsilon} \times \mathbf{\beta}_{i} + \mathbf{\alpha}_{i}) \odot \mathbf{S}_i,
\end{equation}
where $\mathbf{X}_{\mathbf{S}_i} = \mathbf{X} \odot \mathbf{S}_i$. $\mathbf{\beta}_{i}$ and $\mathbf{\alpha}_{i}$ are learnable parameter vectors ($\in \mathcal{R}^{1 \times 1 \times c}$) for the affine transformation, initialized to 1 and 0, respectively. $\mu(\cdot)$ and $\sigma(\cdot)$ compute the mean and standard deviation from the instance, respectively.

The final output of the proposed module considers the original input feature maps as
\begin{equation}
AN(\mathbf{X}) = \rho \mathbf{\bar{X}} + \mathbf{X},
\end{equation}
where $\rho$ is a learnable scalar initialized as 0. Such a residual learning scheme smooths the learning curve by gradually paying more attention to regional normalization.

\begin{figure}[!t]
	\begin{center}
		\centering
		\begin{tabular}{cccc}
			(\textcolor{blue}{-0.08}, \textcolor{red}{4.77}) & (\textcolor{blue}{0.00}, \textcolor{red}{0.35}) & (\textcolor{blue}{0.01}, \textcolor{red}{0.44}) & (\textcolor{blue}{-0.01}, \textcolor{red}{1.09}) \\
			\multicolumn{4}{c}{\includegraphics[width=0.98\linewidth]{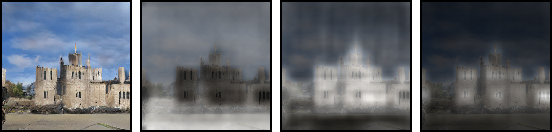}} \\
			(\textcolor{blue}{-0.17}, \textcolor{red}{5.40}) & (\textcolor{blue}{0.01}, \textcolor{red}{0.41}) & (\textcolor{blue}{0.03}, \textcolor{red}{0.71}) & (\textcolor{blue}{-0.02}, \textcolor{red}{0.81}) \\
			\multicolumn{4}{c}{\includegraphics[width=0.98\linewidth]{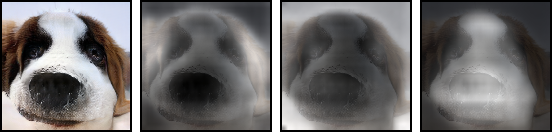}}\\
			\hspace{0.04\columnwidth}(a) & \hspace{0.02\columnwidth}(b) & \hspace{0.04\columnwidth}(c) & \hspace{0.02\columnwidth}(d) \\
		\end{tabular}
	\end{center}
	\vspace{-0.05in}
	\caption{Illustration of how the feature statistics of the feature maps are affected by their computed regions. (a) Generation result. (b-d) Learned attention maps of our method on ImageNet dataset \cite{deng2009imagenet}. Their above tuples indicate the computed \textcolor{blue}{mean} and \textcolor{red}{standard deviation} on the corresponding $32 \times 32$ feature maps. The statistics are calculated on the whole region of (a) and are only processed on the highlighted regions of (b-d).}
	\label{fig_feat_statistics}
	\vspace{-0.05in}
\end{figure}

\begin{figure*}[!t]
	\begin{center}
		\centering
		\begin{tabular}{cccccccc}
			\multicolumn{8}{c}{\includegraphics[width=0.98\linewidth]{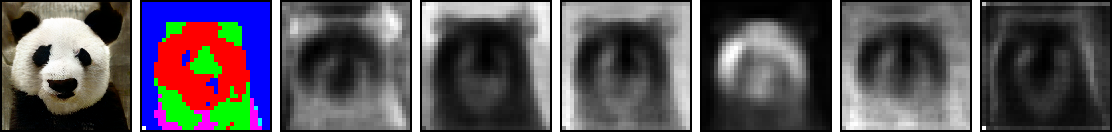}}\\
			\multicolumn{8}{c}{\includegraphics[width=0.98\linewidth]{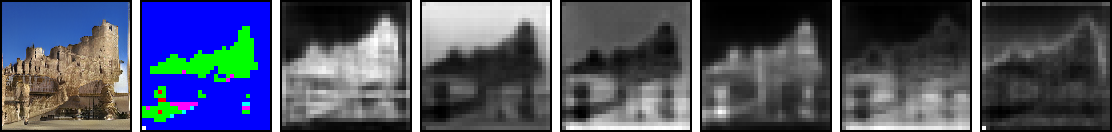}}\\
			\hspace{0.1\columnwidth}(a) & \hspace{0.16\columnwidth}(b) & \hspace{0.16\columnwidth}(c) & \hspace{0.16\columnwidth}(d) & \hspace{0.18\columnwidth}(e) & \hspace{0.16\columnwidth}(f) & \hspace{0.16\columnwidth}(g) & \hspace{0.08\columnwidth}(h)\\
		\end{tabular}
	\end{center}
	\vspace{-0.1in}
	\caption{Visualization of the learned semantic layout on ImageNet. (a) Class-conditional generation results from our method. (b) Binary-version of the learned semantic layout. (c-h) Attention maps activated by the learned semantic entities. The brighter the activated regions are, the higher correlation they are with the used semantic entity. The resolution of the input feature maps is $32 \times 32$.}
	\label{fig_atten_vis}
	\vspace{-0.05in}
\end{figure*}

\subsection{Analysis} \label{sec_analysis}

\paragraph{Why Self-sampling Regularization Works}  It can adaptively capture semantics from the current feature maps, producing proper semantic entity candidates when partial semantic entities are not well learned. The uniform sampling makes such a process not favor specific types of semantics in the early training stage, when the deep features cannot capture semantics. 

Moreover, such sampling makes the employed entity alternatives change during training. We note that the variation of the activated alternatives for useless entities is crucial for learning of the semantic entities, since it can stimulate the current learned useless entities to capture existing semantics in the input feature maps. It is experimentally validated in our experiments (Sec. \ref{sec_exp}). In short, this strategy regularizes SLL from only learning a single semantic entity and leads to understanding more existing semantics.

\vspace{-0.05in}
\paragraph{The Effectiveness of the Learned Semantic Layout}
The predicted semantic layout indicates regions with high inner coherence in semantics. As shown in Figure \ref{fig_feat_statistics}, standard deviation computed from the areas highlighted by our predicted semantic layouts is much lower than that from the whole intermediate feature maps of our generated image (0.35, 0.44, and 1.09 v.s. 4.77 in the 1st row, and 0.41, 0.71, and 0.81 v.s. 5.40 in the 2nd row). Normalizing these points regionally based on their similarities can better preserve the learned semantics.

As shown in Figure \ref{fig_atten_vis}, the learned semantic entities show their diversities by activating different regions of the feature maps. Note the salient foreground object can be detected as the background part. Some entities focus on parts of the object as these regions are highly correlated with the given label information. As shown in (c) and (f) in the 1st row, they highlight the ear/body and facial regions of a panda, respectively, which contain highly discriminative features for this class.

\begin{figure}[t]
	\centering
	\includegraphics[width=0.8\linewidth]{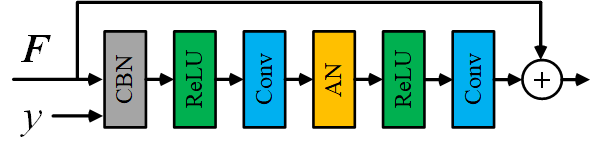}
	\caption{The residual block using attentive normalization.}
	\label{fig:residual-unitv2}
	\vspace{-0.05in}
\end{figure}

\vspace{-0.05in}
\paragraph{Complexity Analysis}
Besides the convolutional computation for generating the intermediate feature maps, the main computation lies on self-sampling and regional normalization. Both of them cost $\mathcal{O}(NHWnC)$, leading to the final $\mathcal{O}(nNHWC)$, where $N$, $H$, $W$, and $C$ denote the batch size, height, width, and the channel number of the input feature maps, respectively. 

AN consumes much less than a self-attention module (with time complexity $\mathcal{O}(N(H^2W^2C+HWC^2))$). It does not have the square term regarding the spatial size of the feature map.

\vspace{-0.05in}
\paragraph{Relation to other Normalizations} Our work is correlated with the existing conditional normalization methods, \eg\ adaptive instance normalization (AdaIN) \cite{huang2017arbitrary} and spatially-adaptive (de)normalization (SPADE) \cite{park2019semantic}. A major difference is that the extra condition (semantic layout) for AN is self-learned from the input features instead of being given as the additional input. Besides, AN treats a spatial portion (indicated by the learned semantic layout) of the features from an image as an instance for normalization.

\section{Applications with Attentive Normalization}
In common practice, AN is placed between the convolutional layer (fully connected layer is not considered because it is computed globally) and the activation layer. To conduct long-range dependency modeling, it should be placed at the relatively large feature maps. Meanwhile, it needs to work on deep layers for the self-sampling regularization. 

Similar to that of \cite{miyato2018cgans}, our proposed AN is incorporated into a residual block \cite{he2016deep} for conditional image generation (shown in Figure \ref{fig:residual-unitv2}).
Since it has a relatively higher complexity than common normalization, we only apply it once in the generative networks, and find it is enough for improving the distant relation as verified in Section \ref{sec_exp}.

In the testing phase, we remove the randomness in the self-sampling branch of AN by switching off this branch with $\mathbf{t}=\mathbf{0}$. Thus, the generation procedure is deterministic only affected by the input.

We integrate AN into two GAN frameworks for class-conditional image generation and generative image inpainting, respectively. The detailed design of the frameworks is given in the supplementary file.

\vspace{-0.05in}
\paragraph{Class-conditional Image Generation}
This task learns to synthesize image distributions by training on the given images. It maps a randomly sampled noise $z$ to an image $x$ via a generator $G$, conditioning on the image label $y$. Similar to that of \cite{miyato2018spectral,zhang2018self}, our generator $G$ is sequentially formed by five residual blocks \cite{he2016deep}, and employs AN in the third residual block (Figure \ref{fig:residual-unitv2}). It outputs $32 \times 32$ feature maps. Also, the discriminator $D$ consists of five residual blocks -- the first one is incorporated with AN.

For the optimization objective, hinge adversarial loss is used to train the generator as
\begin{equation} \label{eq_d}
\mathcal{L}_G = -\text{E}_{z \sim \mathbb{P}_z, y \sim \mathbb{P}_\text{data}}D(G(z,y), y).
\end{equation}
Its corresponding discriminator updating loss is
\begin{equation} \label{eq_g}
\begin{split}
\mathcal{L}_D = &\text{E}_{(x,y) \sim \mathbb{P}_\text{data}}[\text{min}(1-D(x,y))] +\\ &\text{E}_{z \sim \mathbb{P}_z, y \sim \mathbb{P}_\text{data}}[\text{min}(1+D(G(z,y),y))].
\end{split}
\end{equation}

\vspace{-0.15in}
\paragraph{Generative Image Inpainting}
This task takes an incomplete image $\mathbf{C}$ and a mask $\mathbf{M}$ (with missing pixels value 1 and known ones 0) as input and predicts a visually plausible result based on image context. The generated content should be coherent with the given context. Exploiting the known regions to fill the missing ones is crucial for this task. 

Similar to that of \cite{yu2018generative},  we employ a two-stage neural network framework. Both stages utilize an encoder-decoder structure. The AN module is placed in the second stage for exploiting the context to refine the predicted regions.

The learning objective of this task consists of a reconstruction term and an adversarial term as
\begin{equation}
\mathcal{L}_G = \lambda_{rec}||G(\mathbf{C}, \mathbf{M})-\mathbf{Y}||_1  - \lambda_{adv} \text{E}_{\mathbf{\hat{C}} \sim \mathbb{P}_{\mathbf{\hat{C}}}}[D(\mathbf{\hat{C}})],
\end{equation}
where $\mathbf{Y}$ is the corresponding ground truth of $\mathbf{C}$, $\mathbf{\hat{C}}=G(\mathbf{C}, \mathbf{M}) \bigodot \mathbf{M} + \mathbf{Y} \bigodot (\mathbf{1}-\mathbf{M})$, $\mathbb{P}$ denotes data distribution, and $D$ is a discriminator for the adversarial training. $\bigodot$ denotes element-wise multiplication. $\lambda_{rec}$ and $\lambda_{adv}$ are two hyper-parameters for controlling the influence of the reconstruction and adversarial terms.

For adversarial training of the discriminator $D$, WGAN-GP loss \cite{gulrajani2017improved} is adopted as
\begin{equation}
\begin{split}
\mathcal{L}_{D}=&\text{E}_{\mathbf{\hat{C}} \sim \mathbb{P}_\text{data}}[D(\mathbf{\hat{C}})]-\text{E}_{{\mathbf{Y}} \sim \mathbb{P}_\text{data}}[D(\mathbf{Y})]\\&+\lambda_{gp} \text{E}_{\tilde{{\mathbf{C}}}\sim \mathbb{P}_{\tilde{{\mathbf{C}}}}}[(||\nabla_{\tilde{{\mathbf{C}}}}D(\tilde{{\mathbf{C}}})||_2-1)^2],
\end{split}
\end{equation}
where $\tilde{{\mathbf{C}}}=t\mathbf{\hat{C}}+(1-t){\mathbf{Y}}$, $t \in [0, 1]$, and $\lambda_{gp}=10$.


\section{Experimental Results and Analysis} \label{sec_exp}
We evaluate the long-range dependency modeling ability of our AN in the tasks of class-conditional image generation and generative image inpainting. Both tasks rely heavily on distant visual relationship modeling for generating convincing semantic structures for objects and complex scenes. The first task is conducted on ImageNet \cite{deng2009imagenet} (with $128 \times 128$ resolution), while the second one is carried out on Paris Streetview \cite{pathak2016context} (with $256 \times 256$ resolution).

\vspace{-0.05in}
\paragraph{Baselines} Spectral-normalization GAN (SN-GAN) \cite{miyato2018spectral} and self-attention GAN (SA-GAN) \cite{zhang2018self} are adopted as our baselines considering their improvement in class conditional image generation task with popular modular designs. 

BigGAN \cite{brock2018large} and its following work \cite{lucic2019high,zhang2019consistency} are not included since the big model capacity and big batch size are beyond our computation ability. For image inpainting, we take contextual attention (CA) \cite{yu2018generative} as the baseline.

\vspace{-0.05in}
\paragraph{Evaluation Metrics} For quantitative analysis, we adopt Frechet Inception Distance (FID) \cite{heusel2017gans}, intra FID \cite{miyato2018spectral}, and Inception Score (IS) \cite{salimans2016improved} for class conditional image generation task. We employ peak signal-to-noise ratio (PSNR), structural similarity (SSIM), and mean absolute error (MAE) for image inpainting. 

Intra FID gives FID between the generated and real images for a specific class, while FID alone in the following experiments indicates the difference between the synthesized images and real ones with all classes. FID, intra FID, and IS are computed on 50k randomly generated images.

\subsection{Implementation Details}
\paragraph{Class-conditional Image Generation} The Adam optimizer \cite{kingma2014adam} is used. The two-time scale updating scheme \cite{heusel2017gans} is adopted with a $1 \times 10^{-4}$ learning rate for the generator, and a $4 \times 10^{-4}$ learning rate for the discriminator. $\beta_1=0$ and $\beta_2=0.999$. Also, we apply spectral normalization \cite{miyato2018spectral} to both the generator and discriminator to stabilize the training procedure further. All baselines are training with the same batch size 256.

\vspace{-0.15in}
\paragraph{Generative Inpainting} 
To stabilize the training process and generate context-coherent contents, a two-phase training scheme is employed \cite{iizuka2017globally,yu2018generative,wang2018inpainting,,wang2019wide}. In the first training phase, only reconstruction loss is used (by setting $\lambda_{adv}=0$), after the whole training converges. The second phase begins by setting $\lambda_{adv}=1e-3$. In both stages, Adam optimizer is employed with learning rate $=1e-4$, $\beta_1=0.5$ and $\beta_2=0.9$. 

\begin{table}[t]
	\centering
	\caption{Quantitative results of our proposed module on ImageNet with class-conditional generation. SN-GAN* applies spectral normalization to the generator and discriminator, while SN-GAN only applies that to the discriminator. \vspace{0.05in}}
	\label{tb_index_gen_imagenet}
	\small
	\begin{tabular}{c|cccc}
		\hline
		Model & Itr $\times 1K$ & FID $\downarrow$  & Intra FID $\downarrow$ & IS $\uparrow$\\
		\hline
		AC-GAN \cite{odena2017conditional} & / & / & 260.0 & 28.5\\
		SN-GAN \cite{miyato2018cgans} & 1000 & 27.62 & 92.4 & 36.80\\
		SN-GAN* \cite{zhang2018self} & 1000 & 22.96 & / & 42.87\\
		SA-GAN \cite{zhang2018self} & 1000 & 18.65 & 83.7 & $\mathbf{52.52}$\\
		Ours  & $\mathbf{880}$ & $\mathbf{17.84}$ & $\mathbf{83.40}$ & 46.57\\
		\hline
	\end{tabular}
\end{table}

\begin{figure*}[!t]
	\begin{center}
		\centering
		\begin{tabular}{cccc}
			\includegraphics[width=0.24\linewidth]{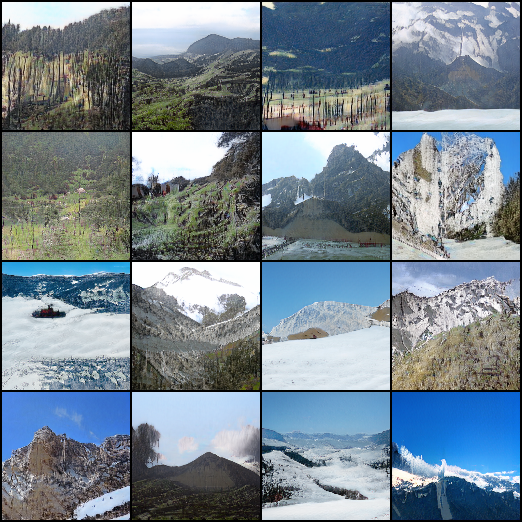} &
			\includegraphics[width=0.24\linewidth]{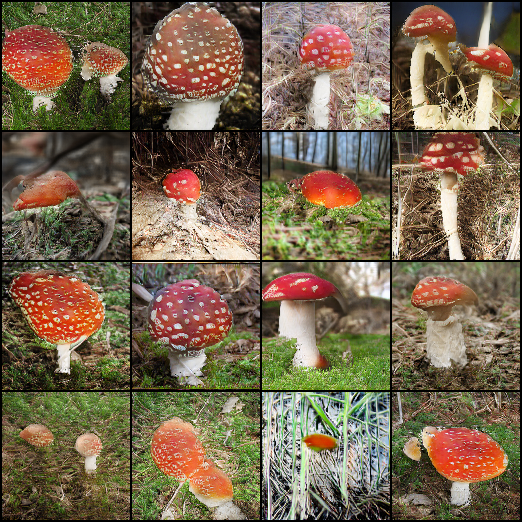} &
			\includegraphics[width=0.24\linewidth]{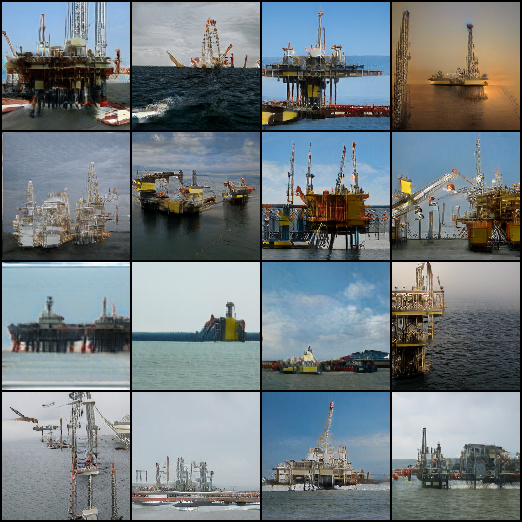} & \includegraphics[width=0.24\linewidth]{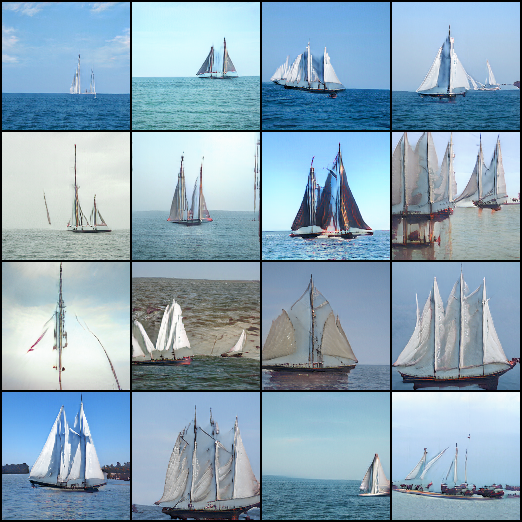}\\
			\hspace{0.00\columnwidth} Alp (970) & \hspace{0.00\columnwidth} Agaric (992) & \hspace{0.00\columnwidth} Drilling platform (540) & \hspace{0.00\columnwidth} Schooner (780) \\
		\end{tabular}
	\end{center}
	\vspace{-0.1in}
	\caption{Randomly generated images ($128 \times 128$) by our model on ImageNet.}
	\label{fig_imagenet_gen}
\end{figure*}

\subsection{Class-conditional Image Generation}

As listed in Table \ref{tb_index_gen_imagenet}, the GAN equipped with our proposed AN module outperforms SN-GAN and SN-GAN* in terms of FID and intra FID. It means our method generates more realistic and diverse visual results compared with the two baselines, validating the effectiveness of AN in this task by capturing the distant relationship. 

Compared with SA-GAN, our method yields lower FID, intra FID, and IS. It shows our module performs comparably to self-attention, which further verifies AN can improve class-conditional image generation performance. About the training iterations to reach convergence, our method costs 880K iterations compared with 1000K by SN-GAN, SN-GAN*, and SA-GAN. Our method has a higher convergence speed in training.

\begin{table}[!t]
	\centering
	\caption{Intra-FID comparison (the lower the better) on typical image classes from ImageNet with class-conditional generation.\vspace{0.05in}}
	\label{tb_index_imagenet_classes}
	\small
	\begin{tabular}{c|ccc}
		\hline
		Class name (label) & SN-GAN \cite{miyato2018cgans} & SA-GAN \cite{zhang2018self} & Ours \\
		\hline
		Stone wall (825) & 49.3 & 57.5 & \textbf{34.16}  \\
		Geyser (974)  & 19.5 & 21.6 & \textbf{13.97}  \\
		Valley (979) & 26.0 & 39.7 & \textbf{22.90}  \\
		Coral fungus (991) & 37.2 & 38.0 & \textbf{24.02}  \\
		\hline
		Indigo hunting (14)  & 66.8 & 53.0 & \textbf{42.54}  \\
		Redshank (141) & 60.1 & 48.9  & \textbf{39.06}  \\
		Saint bernard (247) & 55.3 & \textbf{35.7} & 39.36 \\
		Tiger cat (282) & 90.2 & 88.1 & \textbf{66.65}  \\
		\hline
	\end{tabular}
	\vspace{-0.1in}
\end{table}

Another advantage of our AN is its consistent improvement of generation performance with both relatively simple spatial constraints (\eg\ natural scenes or textures in the first four rows in Table \ref{tb_index_imagenet_classes}) and complex structural relationship (\eg\ objects given in the last four rows in Table \ref{tb_index_imagenet_classes}). 

Table \ref{tb_index_imagenet_classes} shows that our method improves intra FID by a large margin compared with SN-GAN in both cases. It also yields better or comparable intra FID scores compared with SA-GAN. Figure \ref{fig_imagenet_gen} validates that AN well handles textures (alp and agaric) and sensitive structures (drilling platform and schooner) in the visual evaluation. Note that self-attention does not show superiority in the former cases with simple geometrical patterns. 

We observe that our method can produce more diverse patterns on natural scenes or textures. It is because self-attention exerts substantial structural constraints as it uses similar feature points to reconstruct each feature point, which makes the produced features tend to be uniform. Meanwhile, AN enhances spatial relationships regionally, where each region shares the same semantics by normalization. Regional normalization is beneficial to create more diverse patterns compared with the weighted sum of all feature points in the attention mechanism.

\vspace{-0.05in}
\paragraph{Categorical Interpolation}

\begin{figure}[!t]
	\begin{center}
		\centering
		\begin{tabular}{cccccccc}
			\multicolumn{8}{c}{\includegraphics[width=0.98\linewidth]{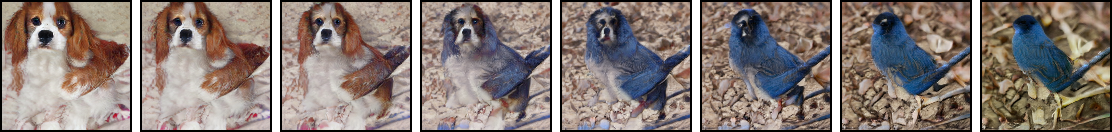}} \\
			\multicolumn{8}{c}{\includegraphics[width=0.98\linewidth]{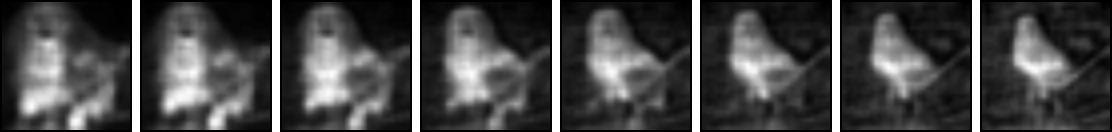}}\\
			\multicolumn{8}{c}{\includegraphics[width=0.98\linewidth]{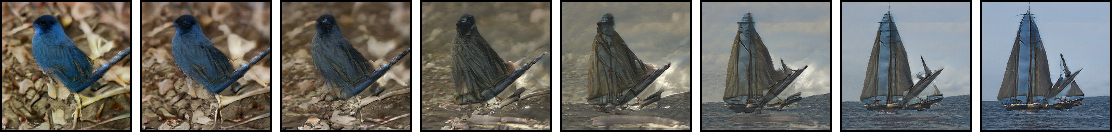}}\\
			\multicolumn{8}{c}{\includegraphics[width=0.98\linewidth]{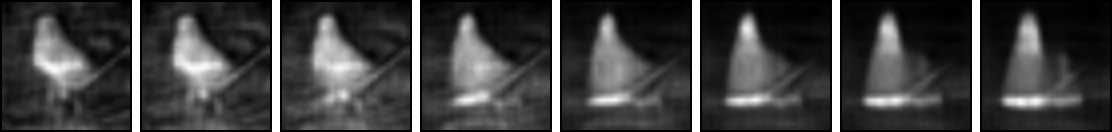}}\\
		\end{tabular}
	\end{center}
	\vspace{-0.05in}
	\caption{Categorical interpolation and intermediate results by our method from Blenheim spaniel (label: 156) to indigo hunting (label: 14), and from indigo hunting to schooner (label: 780) with a fixed noisy signal $z$. 1st and 3rd rows: class-conditional generation results from our method. 2nd and 4th rows: attention maps activated by one semantic entity. The brighter the activated regions are, the higher correlation they are with the used semantic entity.	\vspace{-0.05in}}
	\label{fig_categorical_interp}
\end{figure}

The categorical interpolation of our method can be conducted by the linear combination of the statistics from the used conditional batch normalization with different labels and a fixed input noise $z$ in the generator. Figure \ref{fig_categorical_interp} gives an example. Note that the attention maps given by one semantic entity keep track of the almost-foreground part of the generated image no matter how the foreground changes gradually. It manifests the generality of the learned semantic entities.

\subsection{Applications on Generative Image Inpainting}

\begin{figure*}[!h]
	\begin{center}
		\centering
		\begin{tabular}{cccccc}
			\multicolumn{3}{c}{\includegraphics[width=0.48\linewidth]{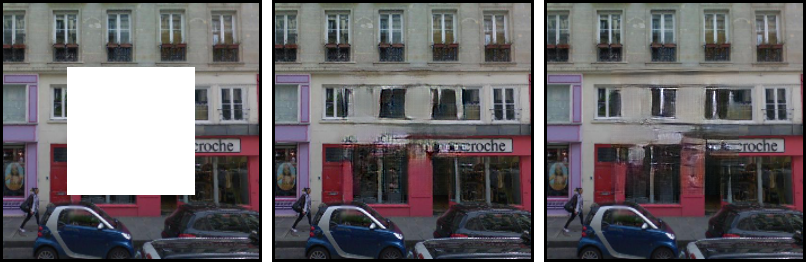}}&
			\multicolumn{3}{c}{\includegraphics[width=0.48\linewidth]{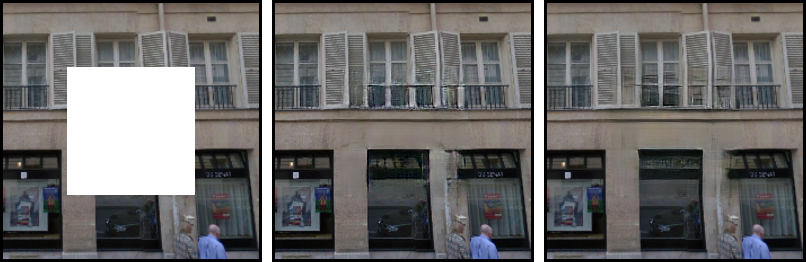}}\\
			\hspace{0.15\columnwidth}(a) & \hspace{0.24\columnwidth}(b) & \hspace{0.12\columnwidth}(c) & \hspace{0.15\columnwidth}(a) & \hspace{0.24\columnwidth}(b) & \hspace{0.12\columnwidth}(c)\\
		\end{tabular}
	\end{center}
	\vspace{-0.1in}
	\caption{Visual comparisons on generative image inpainting on Paris street view. (a) Input images. (b) Results from CA \cite{yu2018generative}. (c)  Ours. More results are given in the supplementary file.	\vspace{-0.05in}}
	\label{fig_inpaint_results}
\end{figure*}

Generative image inpainting relies on long-range interaction and class conditional image generation. A small difference is that the features from context regions are known. 

\begin{table}[!t]
	\centering
	\caption{Quantative comparison on Paris Streetview.\vspace{0.05in}}
	\label{tb_index_paris_streetview}
	\small
	\begin{tabular}{c|ccc}
		\hline
		Method & PSNR (dB) $\uparrow$ & SSIM $\uparrow$ & MAE $\downarrow$ \\
		\hline
		CA \cite{yu2018generative} & 23.78 & 0.8406 & 0.0338\\
		Ours & \textbf{25.09} & \textbf{0.8541} & \textbf{0.0334}\\
		\hline
	\end{tabular}
\end{table}

The inpainting results are given in Figure \ref{fig_inpaint_results}. The baseline equipped with AN yields the most appealing visual performance regarding semantic structure (building facades with windows) and detailed texture. In the quantitative evaluation, our method also performs better than PSRN, SSIM, and MAE, as given in Table \ref{tb_index_paris_streetview}. It, again, validates the effectiveness of AN on enhancing information fusion between cross-spatial regions. 

\begin{table}
	\centering
	\caption{Quantitative results of AN module ablation on ImageNet with class-conditional generation.\vspace{0.05in}}
	\label{tb_index_ablation}
	\small
	\begin{tabular}{c|cc}
		\hline
		Module & IS $\uparrow$ & FID $\downarrow$ \\
		\hline
		Attentive Normalization w BN & 43.92 & 19.59 \\
		Attentive Normalization w/o orthogonal reg & 45.99 & 18.07 \\
		Attentive Normalization w/o SSR & 37.86 & 23.58\\
		Attentive Normalization ($n=8$) & 45.51 & 19.01\\
		Attentive Normalization ($n=16$) & 46.57 & 17.84\\
		Attentive Normalization ($n=32$) & 47.14 & 17.75\\
		\hline
	\end{tabular}
	\vspace{-0.05in}
\end{table}

\subsection{Ablation Studies}
\paragraph{Number $n$ of the Used Semantic Entities} Correlation between feature points is implicitly characterized by the employed semantic entities. Their quantity $n$ controls the fineness of such characterization. The last three rows in Table \ref{tb_index_ablation} show the obvious performance improvement of AN from $n=8$ to $n=16$, while such improvement is relatively marginal from $n=16$ to $n=32$. Considering the trade-off between the effectiveness and efficiency of AN, we choose $n=16$ for experiments in this paper.

\vspace{-0.05in}
\paragraph{Effectiveness of Self-sampling Regularization (SSR)} SSR facilitates the entities in the semantic layout learning (SSL) module to capture meaningful semantics. As mentioned in Sec. \ref{sec_sll}, SSL without SSR tends to produce trivial semantic layouts with only one useful entity (examples are given in the supplementary file). 

In this scenario, regional instance normalization degrades to vanilla instance normalization. Table \ref{tb_index_ablation} shows that our method with SSR yields much lower FID as 17.84 compared with that without it (23.58), where the latter is close to that of SN-GAN* ($22.96$) in Table \ref{tb_index_gen_imagenet}. We suppose the relatively lower performance is caused by the fact that instance normalization does not input the extra label as conditional batch normalization in SN-GAN*.

\vspace{-0.05in}
\paragraph{Choices of Used Normalization in Regional Normalization} Various available forms of normalization \cite{ioffe2015batch,ulyanov2016instance,wu2018group,ba2016layer} can be used here. For simplicity, we only plug and evaluate BN and IN. The lower FID by IN (17.84) compared with that (19.59) by BN shows the relative superiority of IN in this task.

\begin{table}
	\centering
	\caption{Inference time (ms) of our proposed module and self-attention. All fed tensors are with the same batch size 1 and channel number 32. Resolutions are different. `-' stands for evaluation time unmeasurable due to out-of-memory in GPU.\vspace{0.05in}}
	\label{tb_inference_time}
	\scriptsize
	\begin{tabular}{c|cccc}
		\hline
		Module & $128 \! \times \! 128$ & $256 \! \times \! 256$ & $512 \! \times \! 512$ & $1024 \! \times \! 1024$ \\
		\hline
		AN & 0.73 & 2.24 & 9.46 & 37.68\\
		Self-attention & 5.21& 79.42 & - & -\\
		\hline
	\end{tabular}
	\vspace{-0.05in}
\end{table}

\vspace{-0.05in}
\paragraph{The Empirical Evaluation of Computational Efficiency}
The computational efficiency of a neural network module relies on its implementation, software and hardware platform. Here we give the efficiency evaluation of self-attention and our proposed AN (with $n=16$) just for reference. Both of them are programmed with Pytorch 1.1.0, running on the same computational platform with 4 CPUs, 1 TiTAN 2080 GPU, and 32GB memory. 

Table \ref{tb_inference_time} presents that AN performs more efficiently than self-attention concerning both time and GPU memory consumption on the relative large feature maps. Consistent with the complexity analysis in Section \ref{sec_analysis}, the time complexity (empirically) of AN grows linearly with the increase of spatial size, while that of self-attention grows much faster.

\section{Conclusion}
In this paper, we have proposed a novel method to conduct distant relationship modeling in conditional image generation through normalization. It offers a new perspective to characterize the correlation between neural activities beyond the scope limit. Our proposed normalization module is composed of semantic layout learning and regional normalization. The learned semantic layout is sufficient for the regional normalization to preserve and enhance the semantic correspondence learned from the generator. We will explore its usage and possible variants in other tasks (\eg\ classification and semantic segmentation) in future work.

{\small
\bibliographystyle{ieee_fullname}
\bibliography{egbib-norm}

\begin{thebibliography}{10}\itemsep=-1pt

\bibitem{wang2019wide}
Wide-context semantic image extrapolation.

\bibitem{antoniou2017data}
Antreas Antoniou, Amos Storkey, and Harrison Edwards.
\newblock Data augmentation generative adversarial networks.
\newblock {\em arXiv preprint arXiv:1711.04340}, 2017.

\bibitem{arjovsky2017wasserstein}
Martin Arjovsky, Soumith Chintala, and L{\'e}on Bottou.
\newblock Wasserstein generative adversarial networks.
\newblock In {\em ICML}, pages 214--223, 2017.

\bibitem{ba2016layer}
Jimmy~Lei Ba, Jamie~Ryan Kiros, and Geoffrey~E Hinton.
\newblock Layer normalization.
\newblock {\em arXiv preprint arXiv:1607.06450}, 2016.

\bibitem{brock2018large}
Andrew Brock, Jeff Donahue, and Karen Simonyan.
\newblock Large scale gan training for high fidelity natural image synthesis.
\newblock {\em arXiv preprint arXiv:1809.11096}, 2018.

\bibitem{caron2018deep}
Mathilde Caron, Piotr Bojanowski, Armand Joulin, and Matthijs Douze.
\newblock Deep clustering for unsupervised learning of visual features.
\newblock In {\em ECCV}, pages 132--149, 2018.

\bibitem{deng2009imagenet}
Jia Deng, Wei Dong, Richard Socher, Li-Jia Li, Kai Li, and Li Fei-Fei.
\newblock Imagenet: A large-scale hierarchical image database.
\newblock In {\em CVPR}, pages 248--255, 2009.

\bibitem{fu2019dual}
Jun Fu, Jing Liu, Haijie Tian, Yong Li, Yongjun Bao, Zhiwei Fang, and Hanqing
  Lu.
\newblock Dual attention network for scene segmentation.
\newblock In {\em Proceedings of the IEEE Conference on Computer Vision and
  Pattern Recognition}, pages 3146--3154, 2019.

\bibitem{goodfellow2014generative}
Ian Goodfellow, Jean Pouget-Abadie, Mehdi Mirza, Bing Xu, David Warde-Farley,
  Sherjil Ozair, Aaron Courville, and Yoshua Bengio.
\newblock Generative adversarial nets.
\newblock In {\em NeurIPS}, pages 2672--2680, 2014.

\bibitem{greff2017neural}
Klaus Greff, Sjoerd van Steenkiste, and J{\"u}rgen Schmidhuber.
\newblock Neural expectation maximization.
\newblock In {\em NeurIPS}, pages 6691--6701, 2017.

\bibitem{gulrajani2017improved}
Ishaan Gulrajani, Faruk Ahmed, Martin Arjovsky, Vincent Dumoulin, and Aaron~C
  Courville.
\newblock Improved training of wasserstein gans.
\newblock In {\em NeurIPS}, pages 5769--5779, 2017.

\bibitem{he2016deep}
Kaiming He, Xiangyu Zhang, Shaoqing Ren, and Jian Sun.
\newblock Deep residual learning for image recognition.
\newblock In {\em CVPR}, pages 770--778, 2016.

\bibitem{heusel2017gans}
Martin Heusel, Hubert Ramsauer, Thomas Unterthiner, Bernhard Nessler, and Sepp
  Hochreiter.
\newblock Gans trained by a two time-scale update rule converge to a local nash
  equilibrium.
\newblock In {\em NeurIPS}, pages 6626--6637, 2017.

\bibitem{huang2019interlaced}
Lang Huang, Yuhui Yuan, Jianyuan Guo, Chao Zhang, Xilin Chen, and Jingdong
  Wang.
\newblock Interlaced sparse self-attention for semantic segmentation.
\newblock {\em arXiv preprint arXiv:1907.12273}, 2019.

\bibitem{huang2017arbitrary}
Xun Huang and Serge Belongie.
\newblock Arbitrary style transfer in real-time with adaptive instance
  normalization.
\newblock In {\em ICCV}, pages 1501--1510, 2017.

\bibitem{huang2019ccnet}
Zilong Huang, Xinggang Wang, Lichao Huang, Chang Huang, Yunchao Wei, and Wenyu
  Liu.
\newblock Ccnet: Criss-cross attention for semantic segmentation.
\newblock In {\em Proceedings of the IEEE International Conference on Computer
  Vision}, pages 603--612, 2019.

\bibitem{iizuka2017globally}
Satoshi Iizuka, Edgar Simo-Serra, and Hiroshi Ishikawa.
\newblock Globally and locally consistent image completion.
\newblock {\em TOG}, 36(4):107, 2017.

\bibitem{ioffe2015batch}
Sergey Ioffe and Christian Szegedy.
\newblock Batch normalization: Accelerating deep network training by reducing
  internal covariate shift.
\newblock {\em arXiv preprint arXiv:1502.03167}, 2015.

\bibitem{jolicoeur2018relativistic}
Alexia Jolicoeur-Martineau.
\newblock The relativistic discriminator: a key element missing from standard
  gan.
\newblock {\em arXiv preprint arXiv:1807.00734}, 2018.

\bibitem{karras2017progressive}
Tero Karras, Timo Aila, Samuli Laine, and Jaakko Lehtinen.
\newblock Progressive growing of gans for improved quality, stability, and
  variation.
\newblock {\em arXiv preprint arXiv:1710.10196}, 2017.

\bibitem{karras2018style}
Tero Karras, Samuli Laine, and Timo Aila.
\newblock A style-based generator architecture for generative adversarial
  networks.
\newblock {\em arXiv preprint arXiv:1812.04948}, 2018.

\bibitem{kingma2014adam}
Diederik~P Kingma and Jimmy Ba.
\newblock Adam: A method for stochastic optimization.
\newblock {\em arXiv preprint arXiv:1412.6980}, 2014.

\bibitem{Quoc2012unsupervised}
Quoc~V. Le, Marc'Aurelio Ranzato, Rajat Monga, Matthieu Devin, Greg Corrado,
  Kai Chen, Jeffrey Dean, and Andrew~Y. Ng.
\newblock Building high-level features using large scale unsupervised learning.
\newblock In {\em ICML}, 2012.

\bibitem{lucic2019high}
Mario Lucic, Michael Tschannen, Marvin Ritter, Xiaohua Zhai, Olivier Bachem,
  and Sylvain Gelly.
\newblock High-fidelity image generation with fewer labels.
\newblock {\em arXiv preprint arXiv:1903.02271}, 2019.

\bibitem{mao2017least}
Xudong Mao, Qing Li, Haoran Xie, Raymond~YK Lau, Zhen Wang, and Stephen~Paul
  Smolley.
\newblock Least squares generative adversarial networks.
\newblock In {\em ICCV}, pages 2813--2821, 2017.

\bibitem{miyato2018spectral}
Takeru Miyato, Toshiki Kataoka, Masanori Koyama, and Yuichi Yoshida.
\newblock Spectral normalization for generative adversarial networks.
\newblock {\em arXiv preprint arXiv:1802.05957}, 2018.

\bibitem{miyato2018cgans}
Takeru Miyato and Masanori Koyama.
\newblock cgans with projection discriminator.
\newblock {\em arXiv preprint arXiv:1802.05637}, 2018.

\bibitem{odena2017conditional}
Augustus Odena, Christopher Olah, and Jonathon Shlens.
\newblock Conditional image synthesis with auxiliary classifier gans.
\newblock In {\em ICML}, pages 2642--2651. JMLR. org, 2017.

\bibitem{park2019semantic}
Taesung Park, Ming-Yu Liu, Ting-Chun Wang, and Jun-Yan Zhu.
\newblock Semantic image synthesis with spatially-adaptive normalization.
\newblock In {\em CVPR}, pages 2337--2346, 2019.

\bibitem{pathak2016context}
Deepak Pathak, Philipp Krahenbuhl, Jeff Donahue, Trevor Darrell, and Alexei~A
  Efros.
\newblock Context encoders: Feature learning by inpainting.
\newblock In {\em CVPR}, pages 2536--2544, 2016.

\bibitem{radford2015unsupervised}
Alec Radford, Luke Metz, and Soumith Chintala.
\newblock Unsupervised representation learning with deep convolutional
  generative adversarial networks.
\newblock {\em arXiv preprint arXiv:1511.06434}, 2015.

\bibitem{salimans2016improved}
Tim Salimans, Ian Goodfellow, Wojciech Zaremba, Vicki Cheung, Alec Radford, and
  Xi Chen.
\newblock Improved techniques for training gans.
\newblock In {\em NeurIPS}, pages 2234--2242, 2016.

\bibitem{ulyanov2016instance}
Dmitry Ulyanov, Andrea Vedaldi, and Victor Lempitsky.
\newblock Instance normalization: The missing ingredient for fast stylization.
\newblock {\em arXiv preprint arXiv:1607.08022}, 2016.

\bibitem{wang2018inpainting}
Yi Wang, Xin Tao, Xiaojuan Qi, Xiaoyong Shen, and Jiaya Jia.
\newblock Image inpainting via generative multi-column convolutional neural
  networks.
\newblock In {\em NeurIPS}, 2018.

\bibitem{wu2018group}
Yuxin Wu and Kaiming He.
\newblock Group normalization.
\newblock In {\em ECCV}, pages 3--19, 2018.

\bibitem{xu2005maximum}
Linli Xu, James Neufeld, Bryce Larson, and Dale Schuurmans.
\newblock Maximum margin clustering.
\newblock In {\em NeurIPS}, pages 1537--1544, 2005.

\bibitem{xu2018attngan}
Tao Xu, Pengchuan Zhang, Qiuyuan Huang, Han Zhang, Zhe Gan, Xiaolei Huang, and
  Xiaodong He.
\newblock Attngan: Fine-grained text to image generation with attentional
  generative adversarial networks.
\newblock In {\em CVPR}, pages 1316--1324, 2018.

\bibitem{yu2018generative}
Jiahui Yu, Zhe Lin, Jimei Yang, Xiaohui Shen, Xin Lu, and Thomas~S Huang.
\newblock Generative image inpainting with contextual attention.
\newblock {\em arXiv preprint arXiv:1801.07892}, 2018.

\bibitem{zhang2018self}
Han Zhang, Ian Goodfellow, Dimitris Metaxas, and Augustus Odena.
\newblock Self-attention generative adversarial networks.
\newblock {\em arXiv preprint arXiv:1805.08318}, 2018.

\bibitem{zhang2019consistency}
Han Zhang, Zizhao Zhang, Augustus Odena, and Honglak Lee.
\newblock Consistency regularization for generative adversarial networks.
\newblock {\em arXiv preprint arXiv:1910.12027}, 2019.

\end{thebibliography}
}

\end{document}